\documentclass[runningheads]{llncs}
\usepackage[T1]{fontenc}
\usepackage{graphicx}
\usepackage{booktabs}

\usepackage{graphicx}
\usepackage{amsfonts,amsmath}
\usepackage{algorithm,algpseudocode}
\usepackage{mathrsfs}
\usepackage{nicefrac}
\usepackage{subcaption}
\usepackage{booktabs}
\usepackage{multirow}
\usepackage{xcolor}
\usepackage{comment}

\usepackage{hyperref}

\usepackage[misc]{ifsym}

\usepackage{mwe}

\usepackage[misc]{ifsym}

\begin{document}

\title{EPIC: \textbf{E}nhancing \textbf{P}rivacy through \textbf{I}terative \textbf{C}ollaboration} 

\author{Prakash Chourasia\inst{1} \and
Heramb Lonkar\inst{1} \and
Sarwan Ali \inst{1}
\and Murray Patterson \inst{1}
}
\authorrunning{P. Chourasia et al.}
%
\institute{Georgia State University, Atlanta GA 30303, USA \\
\email{pchourasia1@student.gsu.edu, hlonkar1@student.gsu.edu, sali85@student.gsu.edu, mpatterson30@gsu.edu}}

\maketitle              

\begin{abstract}
Advancements in genomics technology lead to a rising volume of viral (e.g., SARS-CoV-2) sequence data, resulting in increased usage of machine learning (ML) in bioinformatics.
Traditional ML techniques require centralized data collection and processing, posing challenges in realistic healthcare scenarios. Additionally, privacy, ownership, and stringent regulation issues exist when pooling medical data into centralized storage to train a powerful deep learning (DL) model. The Federated learning (FL) approach overcomes such issues by setting up a central aggregator server and a shared global model. It also facilitates data privacy by extracting knowledge while keeping the actual data private. 
This work proposes a cutting-edge Privacy enhancement through Iterative Collaboration (EPIC) architecture. The network is divided and distributed between local and centralized servers. We demonstrate the EPIC approach to resolve a supervised classification problem to estimate SARS-CoV-2 genomic sequence data lineage without explicitly transferring raw sequence data. We aim to create a universal decentralized optimization framework that allows various data holders to work together and converge to a single predictive model. The findings demonstrate that privacy-preserving strategies can be successfully used with aggregation approaches without materially altering the degree of learning convergence. Finally, we highlight a few potential issues and prospects for study in FL-based approaches to healthcare applications.

\keywords{Federated Learning \and Classification \and Genomic Sequencing \and SARS-CoV-2 \and Spike Protein.}

\end{abstract}


\section{Introduction}
\label{sec_introduction}
The severe acute respiratory syndrome coronavirus 2 (SARS-CoV-2)
outbreak (aka  COVID-19) has posed a serious threat to society at large. This virus is highly contagious and has spread quickly throughout the world~\cite{li2020coronavirus}. 
The COVID-19 pandemic 
has had an overwhelming impact, with more than 676.6 Million cases and 6.8 Million deaths worldwide as of May 2023~\footnote{\url{https://www.who.int/en/activities/tracking-SARS-CoV-2-variants/}}. 
An enormous amount of genomic sequence data is produced as the COVID-19 disease spreads around the world. 
Analyzing and processing millions of genomic sequences is challenging and raises privacy concerns, particularly for developing countries. Economic and political circumstances may discourage them from sharing accurate data. The COVID-19 pandemic has brought attention to the necessity for a cooperative international strategy for using the enormous amount of sequence data~\cite{de2022variant} to comprehend the evolutionary dynamics of viral lineages. 

It is customary to infer evolutionary relationships between sequences, with most research focusing on genomic surveillance within specific geolocations~\cite{frampton2021genomic,taboada2021genetic,benslimane2021one,voloch2021genomic}. However, it is important to acknowledge that pandemics transcend borders, necessitating coordinated strategies.
We must recognize that a pandemic knows no boundaries and thus enhanced coordinated strategies are required. There are several initiatives for the classification of viral clades and lineages, including the Phylogenetic Assignment of Named Global Outbreak (PANGO) Lineages~\cite{hadfield2018nextstrain}, Nextstrain~\cite{hadfield2018nextstrain}, and Global Initiative on Sharing Avian Influenza Data (GISAID) database~\cite{shu2017gisaid}.
However, the pandemic response policy has pushed governments across the globe to rethink publicizing the actual stats and honest reports on the spread. Providing data privacy and extracting knowledge will encourage the participation of more nations, providing better insights into pandemic-like situations and genomic surveillance.

\begin{figure}[h!]
\centering
  \centering
  \includegraphics[scale=0.45]{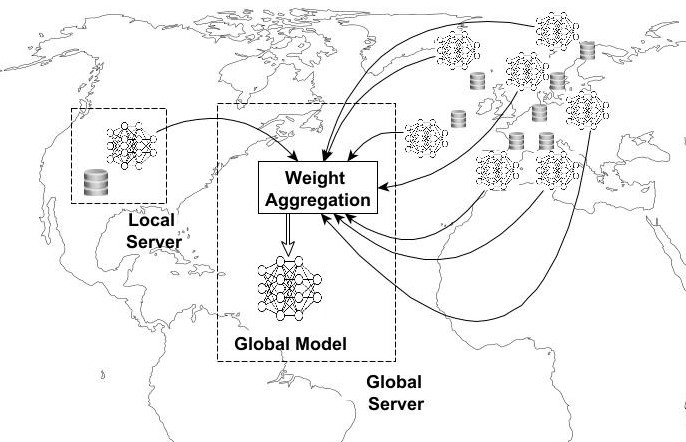}
\caption{
Proposed FL Model Overview (map is taken from~\cite{dmaps_website}).
}
\label{fig_FL_Model}
\end{figure}

FL is a distributed processing system with ML, data privacy, personalized models, and AI at its core. It is a method for training AI models on many data points without transferring or revealing the data from their original place. While FL applies to many fields, it is now being considered for cross-institutional healthcare research~\cite{rieke2020future,yang2019federated}. 
The advantages~\cite{mcmahan2017communication} of using the FL-based approach are: 1) By storing data on local premises, it offers data privacy. 2) It addresses the latency issues since the central server does not require heavy data transfer.
As shown in Figure~\ref{fig_FL_Model}, local models are initially trained locally utilizing the proprietary data of their databases. Next a \emph{global model} is trained using \emph{FL}. A central server referred to as \emph{Global server} is where the global model is stored. \emph{Aggregation function} such as FedAvg~\cite{mcmahan2017communication} aggregates the model parameters from the local models and is sent onto the global server. 
The global model learns from local models through model weight aggregation, enabling continuous learning over time. FL ensures data privacy, overcomes ownership constraints, and generates a generalized intelligent model.

This study demonstrates an FL-based method we call EPIC for SARS-CoV-2 lineage classification using a set of spike sequences. On spike sequence data, we evaluate the performance of our suggested EPIC strategy against expensive baseline techniques, other conventional state-of-the-art (SOTA) methods, and centralized deep learning models. 
We demonstrate the data privacy issue is getting resolved along with an option of getting a better-personalized model for geographical coverage regions for different data sets. 
We see the use of EPIC as a way for authorities and governments to facilitate various privacy concerns while simultaneously extracting the knowledge from private (local) datasets from other nations (private dataset) for a tailored model catered to resolving public health problems and developing policies in a particular context (e.g., state, country, geographical region). 
To perform multi-class classification, we pulled from GISAID the sequences of the $699327$ spike proteins, their lineage data, and other attributes such as location. Our dataset's $5$ distinct lineages came from $8$ different countries. Our contributions are as follows:

\begin{enumerate}


\item We propose EPIC, a scalable and computationally efficient FL model for coronavirus spike sequence classification.

\item In the proposed method, only output weights are communicated from local models to the global model, by refraining from the transfer of actual data we ensure data privacy.

\item We compare our proposed EPIC models with state-of-the-art (SOTA) approaches and assess the pros and cons of our method.

\item We show that the EPIC protects data privacy and helps nations access customized models. These models perform better in terms of the immediate local situation, pick up on patterns, and help to better prepare for a possible spread.
\end{enumerate}

The manuscript is structured as follows: Section \ref{sec_related_work} discusses previous work on FL-based approaches. Section \ref{sec_proposed_approach} presents a proposed FL-based (EPIC) algorithm for sequence classification. Section \ref{sec_experimental_setup} describes the experimental setup and dataset. Section \ref{sec_results_discussion} presents the results, followed by a conclusion in Section \ref{sec_conclusion}.

\section{Related Work}
\label{sec_related_work}

This section reviews the previous works for sequence analysis and the use of ML. We also discuss FL in healthcare and other application domains. We discuss the recent developments and modifications to these methods.
Sequence analysis is vital for understanding the spread of viruses during pandemics~\cite{saravanan2022role,tayebi2021robust,ali2023benchmarking}.
Several methods and tools have been proposed to classify and detect mutations in sequences~\cite{zhang2017viral,chourasia2023reads2vec,murad2023exploring}. 
In a recent study, researchers used phylogenetic reconstruction to identify genomic clades in the Brazil region. ~\cite{voloch2021genomic}. The sheer volume of data involved, however, may limit the effectiveness of established approaches to examining massive datasets.
When compared to conventional sequence lineage prediction techniques, machine learning, and deep learning techniques have several benefits~\cite{atr_B.1.177.21_bik}. They can analyze huge amounts of sequence data, detect patterns and relationships, and generate more precise lineage predictions. Numerous research has employed deep learning and machine learning techniques to assess sequences~\cite{tsai2008exploiting,pirkl2018single,magge2018bi,wang2019predicting}. These models have shown promising results in detecting mutations in viruses and classifying their strains~\cite{anzi2022evaluation}. In order to address the challenges of the huge data set and high dimensionality, several embedding methods have been proposed~\cite{ali2023pssm2vec,ali2023benchmarking,chourasia2022clustering,taslim2023hashing2vec,ali2023anderson}. 
Although machine learning and deep learning methods improve accuracy and scalability in analyzing large datasets, they often compromise data privacy and fairness accountability~\cite{liu2020experiments,pang2021collaborative}, which can undermine authorities' confidence in participating in knowledge sharing.
 Authors in~\cite{chourasia2023efficient} proposed a federated learning-based method for sequence classification, but this method has no iteration. 
 Similarly, authors in~\cite{pang2021collaborative} proposed capturing temporal contexts in historical infection data for COVID-19. 
 DPFact~\cite{ma2019privacy} came up with a privacy-preserving collaborative tensor factorization method for computational phenotyping using electronic health records.
 Vallum-Med~\cite{peterson2020vallum} presented an application of Vallum for the protection of medical patients personal data.
 Federated learning has been successfully used in the image classification domain~\cite{kumar2022neuron,abdul2021covid,zhang2021dynamic,chen2021bridging,li2020review,ahmed2020active,mcmahan2017communication}. Still, it has not been well-explored for biological sequence analysis. 
 Apart from sequence classification, federated learning has also been explored for handling time-series data~\cite{brophy2021estimation} and improving the performance of edge devices in the edge computing domain~\cite{mills2021multi}. 


\section{Proposed Approach}
\label{sec_proposed_approach}
The general basic architecture design and the suggested methodology are outlined in this section before we get into the specifics of the algorithm and workflow.
This section outlines the suggested federated-based approach for classifying coronavirus lineages using spike protein sequences. We first provide a general overview of the proposed approach, and then we delve into the specifics of the algorithm and procedure.

\subsection{Architecture and Models}
\label{arch_and_models}
This section starts by explaining central baseline models followed by the description of the proposed EPIC model.
\subsubsection{Wasserstein Distance Guided Representation Learning (WDGRL)~\cite{shen2018wasserstein}}

WDGRL is a recently proposed Wasserstein GAN-inspired method that can be used in order to reduce domain discrepancy for domain adaptation.
The feature extractor network is then optimized by WDGRL in an adversarial way to reduce the estimated Wasserstein distance using a neural network.
We use WDGRL because it is well-suited here since we can see the class imbalance in the target label for our dataset. 
Due to WDGRL being based on a neural network, its requirement for training data can be expensive.

\subsubsection{Poincaré Embedding~\cite{nickel2017poincare}}
Using the unique characteristics of hyperbolic space, the Hyperbolic Embedding method converts high-dimensional biological data into a lower-dimensional representation. In particular, instead of depending on conventional Euclidean geometry, this method computes pairwise distances between data points in hyperbolic space using Poincaré distance~\cite{nickel2017poincare}. This makes it possible to preserve the rich structural information that biological sequences naturally contain. This approach takes as input a spectrum depending on the total count of amino acids in a given sequence and generates a low-dimensional embedding in hyperbolic space for sequence categorization. Novel sequences can be effectively classified by feeding this reduced representation into traditional machine learning classifiers. 

Subtle patterns and correlations between sequences that may not be visible using conventional Euclidean-based approaches are captured by the ensuing embeddings. The use of hyperbolic geometry in this method has various advantages, one of which is that it ensures the biological sequences retain their intrinsic structure by taking into account the complex interdependencies and interactions among them. Making it possible to calculate pairwise distances between sequences quickly and effectively, even with big datasets. With potential uses in fields including illness detection, customized medicine, and protein function prediction, the Hyperbolic Embedding approach provides a strong tool for evaluating and categorizing biological sequences.



\subsubsection{Centralized Feed Forward Neural Network}
\label{subsect_centralized_FF}

This model uses a centralized neural network architecture trained on aggregated data from multiple countries. It includes dense layers, batch normalization, dropout, and activation functions. Each local model is compiled with the 'Adam' optimizer, 'categorical\_crossentropy' loss function, and accuracy metric. Our global model represents the central baseline in the centralized setting.



\subsubsection{EPIC Model}


Our proposed Federated Learning (FL) framework, EPIC, consists of two core components: Local Models and a Global Model. By distributing the computation and leveraging data from local clients (models from each region/country), we aim to preserve their privacy while extracting valuable insights. This is achieved by having local models process their datasets without sharing actual data with the central server. 
Initially, we establish eight local neural network models, one for each country, using the architecture described in Section~\ref{subsect_centralized_FF}. Each local model is trained on its respective
dataset and outputs weighted information organized by time intervals (monthly in our case). These weights are then transmitted to a central server, where they're aggregated using an averaging technique.
This iterative process continues for each time interval, converging towards well-informed global and local models ultimately.

The Global Model plays a crucial role in EPIC, as it aggregates the weighted information from all local models, providing a comprehensive view of the overall trend. By incorporating this knowledge, we
can make more informed decisions and improve the accuracy of our predictions.
Throughout the iterative process, each local model is trained using its dataset, ensuring that the learned patterns are specific to that region's unique characteristics. Meanwhile, the Global Model
benefits from the collective knowledge shared by all local models, allowing it to capture global trends and relationships that might not be apparent at the local level.
The resulting framework, EPIC, offers a powerful tool for analyzing and modeling complex systems, enabling us to extract valuable insights while preserving the privacy of individual data sources.




\subsection{Algorithm and Workflow}
\label{algo_and_workflow}
\begin{figure*}[ht!]
    \centering
    \includegraphics[scale=0.18]{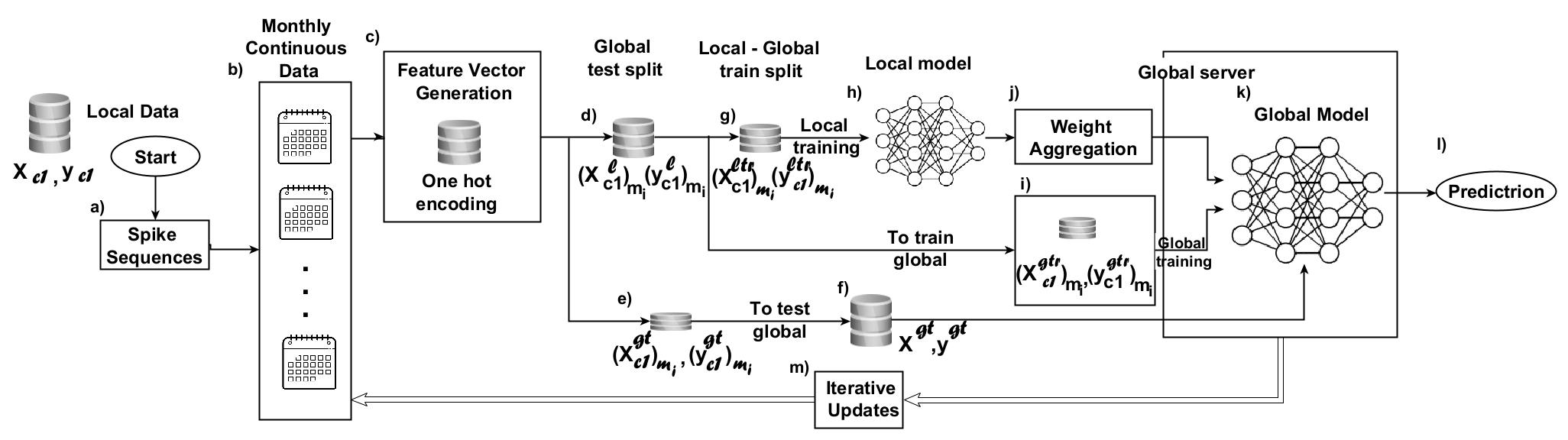}
    \caption{Flow chart for a country collaborating in the FL process}
    \label{fig_FL_Flow_Chart}
\end{figure*}

We construct the numerical feature vectors from the spike sequences and lineages (target labels) by using a one-hot encoding~\cite{kuzmin2020machine} (OHE) on sequences and the labels. The resulting vectors are aligned using the trail zero-padding method. The workflow of EPIC can be seen in Figure~\ref{fig_FL_Flow_Chart} and the algorithmic pseudocode is given in Algorithm~\ref{algo_fedrated}. 

First, we begin with partitioning the targets and labels month-wise 
$X_{m_i}^{total}, y_{m_i}^{total}$ as shown in lines 5 and 6 in Algorithm~\ref{algo_fedrated} 
and Figure~\ref{fig_FL_Flow_Chart}-b. The training cycle iterates on a monthly 
basis. Subsequently, we filter the data country-wise, and some part (30\%) is kept 
aside for global testing data as $(X_{c_j}^{gt})_{m_i}, (y_{c_j}^{gt})_{m_i}$ which is 
further appended to the list to form global testing set $X^{gt}, y^{gt}$ as shown in 
Lines 10, 11, and 12 in Algorithm~\ref{algo_fedrated} also shown in 
Figure~\ref{fig_FL_Flow_Chart}-e and f.
Afterward, the Global training data set is taken out from the left total 
$(X_{c_j}^{gtr})_{m_i},  (y_{c_j}^{gtr})_{m_i}$ as shown in Line 13 of 
Algorithm~\ref{algo_fedrated} and Figure~\ref{fig_FL_Flow_Chart}-i.
This global setting is used for training the Global model later.
Next, we subdivide the total data into the local train 
$ (X_{c_j}^{ltr})_{m_i},(y_{c_j}^{ltr})_{m_i}$ and local test data 
$ (X_{c_j}^{lt})_{m_i}, (y_{c_j}^{lt})_{m_i}$ as shown in Line 14 of 
Algortihm~\ref{algo_fedrated} also shown in Figure~\ref{fig_FL_Flow_Chart}-g and h. 
Here, the local test data from each month is appended to form a list of the country's local test set 
$ X_{c_j}^{lt}, y_{c_j}^{lt}$ as shown in line 15 and 16 of Algortihm~\ref{algo_fedrated}. 
This test set is used to test the final locally trained model for the respective country.
Lines 5 to 16 in the Algorithm~\ref{algo_fedrated} are the data partitioning steps. 
Afterward, after checking whether the local weights $LW_{c_j}$ are empty, we proceed with 
the first round of training a local model.
($l\_model_{c_j}$) using the local training data for the respective month and country 
$(X_{c_j}^{ltr})_{m_i}, (y_{c_j}^{ltr})_{m_i}$. The resulting local weights after 
training are saved in the Local Weights or $LW_{c_j}$. 
For the first round of training the global weights $GW$ are also 
empty. In such instances, we proceed to train the global model ($gb\_model$) using 
the global training data and save them in GlobalWeights $GW$ as shown in 
Algorithm~\ref{algo_fedrated} Line 28. 
For the subsequent iterative training of the local model, if we notice 
the presence of previous local weights $LW_{c_j}$(which also implies the existence of 
previous global weights $GW$), we merge the local weights with the global weights 
for the local model and proceed to train a new local model with merged weights (lines 21 to 23) 
for the next round of training. The newly updated local weights are then saved 
in the Local Weights $LW_{c_j}$. 
For monthly iteration of the global training, results in new global weights being 
saved in the $GW_{i}$ for each iteration $i$. Which is the Aggregation 
of previous $GW$ and all the Local weights from each country $\Sigma LW_{c_j}$. This 
global model is trained on the monthly data $(X_{c_j}^{gtr})_{m_i}, 
(y_{c_j}^{gtr})_{m_i}$, where the model is incorporating the aggregation of weights 
from the local models of all countries for that month of training, plus learning from 
previous iterations.

\begin{algorithm}[h!]
	\caption{EPIC Model Pseudo Code 
 }
\label{algo_fedrated}
	\begin{algorithmic}[1]
 \scriptsize
	\Statex \textbf{Input:} Feature vectors $X$, Labels $y$, $countryList$, and $months$
	\Statex \textbf{Output:} Lineage predictions $V$
	
\State $X^{gt}, y^{gt}$ = $[]$, $[]$
 \State $LocalWeights$ = $[]$ $\backslash$* LW $\rightarrow$ LocalWeights *$\backslash$
        \State $GlobalWeights$ = $[]$ $\backslash$* GW $\rightarrow$ GlobalWeights *$\backslash$

        \For {$m_{i}$ in $months$}
        \State $X_{m_{i}}^{total}$ = TimeWiseSeq($X^{total}$)  \Comment{seq for time $m_i$}
            \State $y_{m_{i}}^{total}$ = TimeWiseLabels($y^{total}$)\Comment{labels for time $m_i$}
	   \For{$c_j$ in $countryList$}

            \State $(X_{c_j}^{total})_{m_i}$ = CntryWiseSeq($X_{m_{i}}^{total}$) \Comment{seq for country j}
	    \State $(y_{c_j}^{total})_{m_i}$ = CntryWiseLbl($y_{m_{i}}^{total}$) 
               \Comment{labels}
     
         \State $(X_{c_j}^{total})_{m_i}, (X_{c_j}^{gt})_{m_i}, (y_{c_j}^{total})_{m_i}, (y_{c_j}^{gt})_{m_i}$ = SplitGlobalTest($(X_{c_j}^{total})_{m_i}$, $(y_{c_j}^{total})_{m_i}$)  
         \State $X^{gt}.$append($(X_{c_j}^{gt})_{m_i}$)
         \State $y^{gt}.$append($(y_{c_j}^{gt})_{m_i}$)

        \State $(X_{c_j}^{total})_{m_i}, (X_{c_j}^{gtr})_{m_i}, (y_{c_j}^{total})_{m_i}, (y_{c_j}^{gtr})_{m_i}$ = SplitLocalGlobalData($(X_{c_j}^{total})_{m_i}$, $(y_{c_j}^{total})_{m_i}$) 
        \State $(X_{c_j}^{ltr})_{m_i}, (X_{c_j}^{lt})_{m_i}, (y_{c_j}^{ltr})_{m_i}, (y_{c_j}^{lt})_{m_i}$ = 
        SplitLocalTrainTest($(X_{c_j}^{total})_{m_i}, (y_{c_j}^{total})_{m_i}$)
        
		\State $X_{c_j}^{lt}.$append($(X_{c_j}^{lt})_{m_i}$)
		\State $y_{c_j}^{lt}.$append($(y_{c_j}^{lt})_{m_i}$)
		
        \If{$LW_{c_j}$ == $None$}
            \State $lc\_model_{c_j}$ = newLocalModel()  \Comment{Local model}

            $\backslash$* LMT $\rightarrow$ Local Model Training *$\backslash$
            
            \State $LW_{c_j}$ = $lc\_model_{c_j}$.LMT($(X_{c_j}^{ltr})_{m_i}, (y_{c_j}^{ltr})_{m_i}$)

        \Else
        \State $w_{c_j}$ = Aggregation($LW_{c_j}$, $GW$)
            \State $lc\_model_{c_j}$ = newLocalModel($w_{c_j}$)
            \State $LW_{c_j}$ = model.LMT($(X_{c_j}^{ltr})_{m_i}, (y_{c_j}^{ltr})_{m_i}$) 
    \EndIf

    \EndFor
	
	    \If{$GW$ == $None$}
            \State gb\_model = newGlobalModel()         \Comment{Local model}

            $\backslash$* GMT $\rightarrow$ Global Model Training *$\backslash$
            
            \State $GW$ = gb\_model.GMT($(X_{c_j}^{gtr})_{m_i}, (y_{c_j}^{gtr})_{m_i}$)

        \Else
        \State $w_{i}$ = Aggregation($LW$, $GW$)
            \State gb\_model = newGlobalModel($w_{i}$)
            \State $GW$ = gb\_model.GMT($(X_{c_j}^{gtr})_{m_i}, (y_{c_j}^{gtr})_{m_i}$) 
    \EndIf
	
    \EndFor
	
	\For{$c_j$} in $countryList$
		\State $pred$ = $lc\_local_{c_j}$.Predict($X_{c_j}^{lt}, y_{c_j}^{lt})$ \Comment{testing local model}
	\EndFor

	\State $pred$ = gb\_model.Predict($X^{gt}, y^{gt})$ \Comment{testing Global model}

    \State return($pred$)
	\end{algorithmic}
\end{algorithm}




\section{Experimental Setup}
\label{sec_experimental_setup}
In this section, we delve into the details of our dataset, comprising spike sequences used for experimental purposes. We provide a comprehensive overview of the number of samples in the dataset. We also explore the baseline model, which serves as a starting point for comparison with our proposed machine-learning techniques. The baseline model is trained on the same dataset and uses the same evaluation metrics as our proposed models. This allows us to evaluate the effectiveness of our proposed approaches relative to a well-established and widely used method in the field.  By examining the baseline model, machine learning techniques, and evaluation metrics, we can gain valuable insights into the performance of our proposed approach and identify areas for further improvement.
To ensure consistency and reproducibility, all experiments were conducted on a powerful Intel(R) Core i5 system running Windows 10 $64$ bit OS, equipped with 32 GB memory.


\subsection{Dataset statistics}
\label{SubSection_Data_Stats}
We used the amino acid sequences corresponding to the
spike protein from the largest known database of SARS-CoV-2
sequences, GISAID. We have the 5 most common
lineages with a total of $699327$ sequences which are used in our experiments are shown in Table~\ref{tbl_variant_information}. The country-wise sequence count distribution is provided in Table~\ref{tbl_data_distribution_CountryWise}.

\begin{table}[ht!]
  \centering
  \begin{tabular}{p{1.2cm}lp{0.9cm}p{2cm} | p{1.2cm}}
    \toprule
    
      \multirow{3}{1.2cm}{Lineage} & \multirow{3}{*}{Region} & \multirow{3}{*}{Labels} &
	\# of Mutations S-gene/Genome &  Num. of sequences\\
      \midrule	\midrule
      B.1.1.7 & UK~\cite{galloway2021emergence} &  Alpha & 8/17 &  593236 \\
      B.1.351  & South Africa~\cite{galloway2021emergence}  &  Beta & 9/21 &   7746\\
      B.1.617.2  & India~\cite{yadav2021neutralization}  &  Delta &  8/17  &   69886 \\
      P.1  &  Brazil~\cite{naveca2021phylogenetic} &  Gamma &  10/21 &   16471\\
      B.1.427   & California~\cite{zhang2021emergence}  & Epsilon  &  3/5 &   11993 \\
      \midrule
      Total & -  & -  & -  &   699327 \\

      \bottomrule
  \end{tabular}
  \caption{Dataset Statistics for different lineages. 
    }
  \label{tbl_variant_information}
\end{table}

\begin{table}[ht!]
  \centering
    \begin{tabular}{cc|cc}
    \toprule
       Region & No. Sequences  &  Region & No. Sequences \\
        \midrule \midrule
England &  245695 &  USA &  190851 \\ 
Germany & 72149 & Denmark &   59353 \\ 
Sweden &   39536 & Scotland &   38054 \\
Netherlands &   27504 & France &   26185 \\ 
\midrule
&  & \textbf{Total} & \textbf{699327} \\
        \bottomrule
    \end{tabular}
  \caption{Country-wise sequences distribution of GISAID dataset.}
  \label{tbl_data_distribution_CountryWise}
\end{table}

\section{Results And Discussion}
\label{sec_results_discussion}
We present the classification results in this part using various evaluation metrics. Table~\ref{Table_NN_Results} compares several neural network-based techniques (baselines) classification results with the suggested EPIC approach. 

We can see that Poincaré Embedding performs comparable to the proposed approach but still is not better with KNN as the best-performing classifier and is behind our proposed model marginally. Moreover, this centralized method needs all data at the server to train the model. Similarly, we get comparable results with WDGRL along with Feed Forward NN. Still, it is a centralized system requiring access to all data in a central server. It violates data privacy and adds overhead for transmitting sensitive and big data to the centralized server. 
Additionally, the results are nearly identical compared to Feed Forwards NN, which uses the same model but operates in a centralized environment without a Federated learning architecture. However, this design necessitates centralized processing. Additionally, it does not guarantee data privacy and requires data transfer to the server.
However, since our goal here is to preserve data privacy while maintaining better predictive performance, we can observe from accuracy, precision, recall, F1 (weighted, Macro), and ROC-AUC results that the proposed EPIC model fulfills that criterion efficiently, and even performs better or comparable when are compared to centralized ML-based approach with similar model architecture. 
Moreover, as the centralized model is trained on the full dataset, while in FL, the local models can access their respective data only, thus, FL distributes the computational load to the clients and preserves and addresses the data privacy issue.

\begin{figure}[h!]
  \centering
  \begin{subfigure}{.45\textwidth}
  \centering
  \includegraphics[scale=0.45]{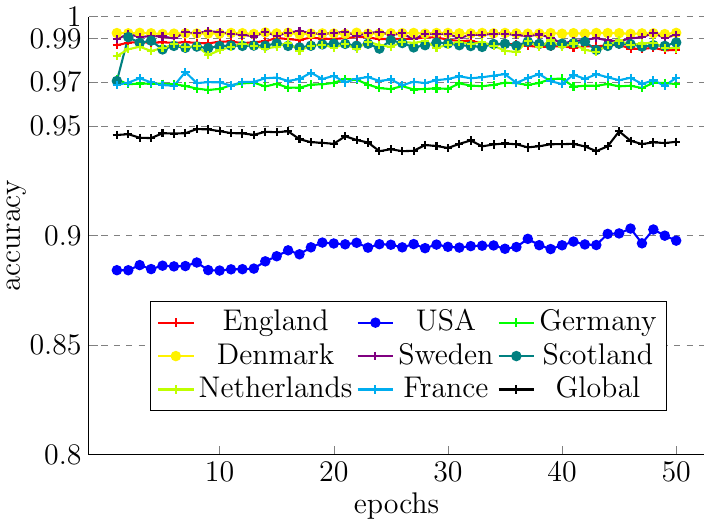}
  \caption{The accuracy curve for different classification models (After training for 6/6 months).}
  \label{fig_local_model_results_accuracy}
  \end{subfigure}%
  \hspace{1em}
  \begin{subfigure}{.45\textwidth}
  \centering
  \includegraphics[scale=0.38]{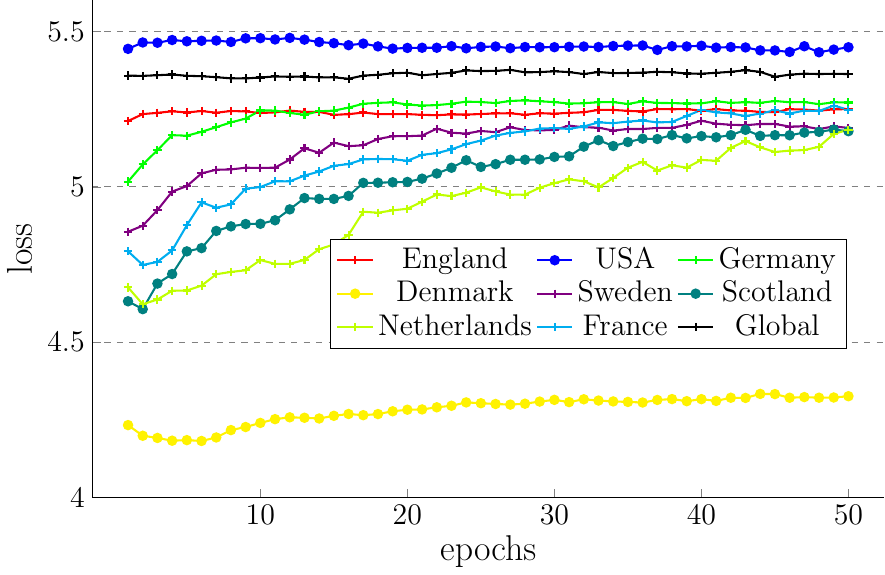}
  \caption{The loss curve for different classification models (After training for 6/6 months)}
  \label{fig_local_model_results_loss}
  \end{subfigure}%
 
  \caption{Training accuracy and loss curve after comp[letting all iterations of different local and Global models while training the model (Local and Global Models). The figure is best seen in color. }
    \label{fig_local_global_model_acc_loss_plots}
\end{figure}

The training accuracy and loss with an increasing number of epochs for all local models and the global model (EPIC) using FLAvgWeight aggregation are shown in Figure~\ref{fig_local_global_model_acc_loss_plots}. The black line plot represents the Global model accuracy and loss for increasing epochs. The accuracy for most local models Figure~\ref{fig_local_global_model_acc_loss_plots}-a is higher than the global (black line). This is expected because we want our local model to perform better on the local dataset. Still, at the same time, the model should learn from other country's datasets to be more robust and perform well on new data. This is achieved by aggregating learning from other local datasets while keeping privacy for these local data (country-specific) using the EPIC.
Similarly, we can see in Figure~\ref{fig_local_global_model_acc_loss_plots}-b the loss is better in the case of the Global model than the local one.



\begin{table*}[h!]
    \centering
    \resizebox{0.99\textwidth}{!}{
    \begin{tabular}{@{\extracolsep{6pt}}p{1cm}ccp{0.9cm}p{0.9cm}p{0.9cm}p{1.1cm}p{1.1cm}p{0.9cm}p{1.4cm}}
    \toprule
        \multirow{2}{*}{Category} & Embedding & \multirow{2}{*}{Method/Model}  & \multirow{2}{*}{Acc. $\uparrow$} & \multirow{2}{*}{Prec. $\uparrow$} & \multirow{2}{*}{Recall $\uparrow$} & \multirow{2}{1.2cm}{F1 (Weig.) $\uparrow$} & \multirow{2}{1.3cm}{F1 (Macro) $\uparrow$} & \multirow{2}{1.2cm}{ROC AUC $\uparrow$} & Train Time (sec.) $\downarrow$
          \\
        \midrule \midrule
         \multirow{14}{1.9cm}{Central Feature Engineering} &  
        \multirow{7}{1.2cm}{WDGRL}
        & SVM & 0.947 & 0.937 & 0.947 & 0.937 & 0.609 & 0.752 & 40606.880 \\
         &  & NB & 0.812 & 0.927 & 0.812 & 0.858 & 0.490 & 0.803 & \textbf{\textcolor{blue}{0.219}} \\
         &  & MLP & 0.960 & 0.958 & 0.960 & 0.956 & 0.771 & 0.829 & 165.978 \\
         &  & KNN & \textbf{\textcolor{blue}{0.968}} & 0.967 & 0.968 & 0.967 & 0.851 & 0.885 & 581.412 \\
         &  & RF & 0.976 & \textbf{\textcolor{blue}{0.976}} & \textbf{\textcolor{blue}{0.976}} & \textbf{\textcolor{blue}{0.974}} & \textbf{\textcolor{blue}{0.885}} & 0.894 & 71.596 \\
         &  & LR & 0.947 & 0.939 & 0.947 & 0.937 & 0.607 & 0.756 & 19.847 \\
         &  & DT & 0.960 & 0.961 & 0.960 & 0.960 & 0.815 & \textbf{\textcolor{blue}{0.898}} & 2.407 \\
        \cmidrule{2-10}
         & \multirow{7}{1cm}{Poincaré embedding}  & 
         SVM & 0.8492 & 0.7212 & 0.8492 & 0.7800 & 0.1837 & 0.5000 & 98.25 \\
         &  & NB & 0.8465 & 0.8214 & 0.8465 & 0.8281 & 0.3278 & 0.5812 & \textbf{\textcolor{blue}{1.29}} \\
         &  & MLP & \textbf{\textcolor{blue}{0.9576}} & 0.9249 & \textbf{\textcolor{blue}{0.9576}} & 0.9409 & 0.5595 & 0.7574 & 144.40 \\
         &  & KNN & 0.8986 & 0.8710 & 0.8986 & 0.8712 & 0.3803 & 0.6021 & 1904.03 \\
         &  & RF & 0.9605 & \textbf{\textcolor{blue}{0.9277}} & 0.9605 & \textbf{\textcolor{blue}{0.9437}} & 0.5739 & 0.7590 & 1064.28 \\
         &  & LR & 0.8492 & 0.7212 & 0.8492 & 0.7800 & 0.1837 & 0.5000 & 89.07 \\
         &  & DT & 0.9206 & 0.9272 & 0.9206 & 0.9239 & \textbf{\textcolor{blue}{0.5785}} & \textbf{\textcolor{blue}{0.7621}} & 125.50 \\
         \cmidrule{2-10} 
          \multirow{2}{1.9cm}{Central NN} & - & CNN & \textbf{\textcolor{blue}{0.990}} & 0.898 & 0.869 & 0.816 & 0.414 & 0.671 & 212670.36 \\
          

          & - & Feed Forward NN & 0.970 & 0.972 & 0.970 & 0.967 & 0.719 & 0.886 & 12383.15 \\
          
          \midrule
          
          \multirow{9}{1.2cm}{EPIC FedAvg} 
          & - & England & 0.934 & 0.917 & 0.934 & 0.921 & 0.537 & 0.777 & 312365.85 \\
          & - & USA & \textbf{0.973} & \textbf{0.969} & \textbf{0.973} & \textbf{0.969} & \textbf{0.711} & \textbf{0.880} & 195311.12 \\
          & - & Germany & 0.962 & 0.963 & 0.962 & 0.960 & 0.690 & 0.878 & 221243.91 \\
          & - & Denmark & 0.840 & 0.768 & 0.840 & 0.802 & 0.199 & 0.533 & 184312.54 \\
          & - & Sweden & 0.937 & 0.900 & 0.937 & 0.918 & 0.562 & 0.752 & 172343.32 \\
          & - & Scotland & 0.926 & 0.884 & 0.926 & 0.904 & 0.375 & 0.663 & 194721.34 \\
          & - & Netherlands & 0.950 & 0.953 & 0.950 & 0.948 & 0.681 & 0.874 & 200865.81 \\
          & - & France & 0.937 & 0.900 & 0.937 & 0.918 & 0.562 & 0.751 & \textbf{160552.97} \\
            \cmidrule{2-10}
          & - & Global & 0.849 & 0.720 & 0.849 & 0.780 & 0.184 & 0.580  & 216937.74\\


          
          \bottomrule
    \end{tabular}
    }
    \caption{Results comparison of different neural network-based methods with proposed EPIC. The Best values for EPIC are shown in bold black, while the best values for the baselines are in blue. 
    }
    \label{Table_NN_Results}
\end{table*}

\section{Conclusion}
\label{sec_conclusion}


In this work, we presented the Enhanced Privacy Iterative Collaboration (EPIC) approach to address data privacy and data availability issues in sequence data processing, especially during pandemics. The feed-forward neural networks are trained locally using our suggested method iteratively, and monthly learning from local models is aggregated to the central server to build a Global Model. Iterative and monthly updates mimic the real-world scenario to integrate learning and collaboration in a timely fashion. We proved the viability of our method and showed that it outperforms conventional centralized deep learning models through trials on a real-world dataset. 
Our findings imply that EPIC can be extremely useful in resolving data availability and privacy problems. In tracing lineages of viruses, the suggested method has produced encouraging results. The global model that results from training on data from several sources (countries) can be generalized better.  Insights that might not be attainable with a single dataset on time can be discovered, and new data can be processed effectively for classification. Future work will explore other deep learning models, like recurrent neural networks, to enhance accuracy. Additionally, we plan to apply this approach to other domains beyond coronavirus to evaluate its effectiveness.
However, there is still potential for improvement in terms of fairness and predictability. To further increase the predicted accuracy, we intend to investigate additional deep learning models such as recurrent neural networks in future work. Aside from coronavirus, we also intend to apply the suggested approach to other domains and assess its effectiveness. 

\bibliographystyle{IEEEtran}
\bibliography{7}

\end{document}